\documentclass[pdftex]{edm_template}
\usepackage{comment}
\usepackage[para]{threeparttable}
\usepackage{multirow}
\usepackage{subcaption}

\begin{document}
\numberofauthors{3} 
%
\author{
%
%
\alignauthor
Zhiyun Ren\\
       \affaddr{Computer Science}\\
       \affaddr{George Mason University}\\
       \affaddr{4400 University Drive,}\\
       \affaddr{Fairfax, VA 22030}\\
       \email{zren4@gmu.edu}
\alignauthor
Xia Ning\\
       \affaddr{Computer \& Information Science}\\
       \affaddr{Indiana University - Purdue University Indianapolis}\\
       \affaddr{420 University Blvd, Indianapolis, IN 46202}\\
       \email{xning@cs.iupui.edu}
\alignauthor
Huzefa Rangwala\\
       \affaddr{Computer Science}\\
       \affaddr{George Mason University}\\
       \affaddr{4400 University Drive,}\\
       \affaddr{Fairfax, VA 22030}\\
       \email{rangwala@cs.gmu.edu}
}

\title{Grade Prediction with Temporal Course-wise Influence}



\maketitle
\begin{abstract}
There is a critical need to develop new educational
  technology applications 
that analyze the data collected by  universities to ensure that 
students graduate in a timely fashion (4 to 6 years); and they are well prepared 
for jobs in their respective fields 
of study. In this paper, we
present a novel approach for analyzing historical educational records 
from a large, public university to perform next-term grade prediction; i.e., to estimate
the grades that a student will get in a course that he/she will enroll in the next term.
Accurate next-term grade prediction holds the promise for better student degree planning, personalized 
advising and automated interventions to ensure that students stay on track in their chosen degree program and graduate 
on time. 
We present a factorization-based approach  called 
Matrix Factorization with Temporal Course-wise Influence
that incorporates  course-wise influence effects and temporal 
effects for grade prediction. 
In this model, students and courses are represented in a latent ``knowledge'' space. The grade of a
student on a course is modeled as the similarity of their latent representation in the 
``knowledge'' space. Course-wise influence is considered as an additional factor in the grade prediction.
Our experimental 
results show that the proposed method outperforms several baseline approaches and infer meaningful patterns between pairs of courses within academic programs.
\end{abstract}


\section{Introduction}








Data analytics  is at 
the forefront of innovation in 
several of today's popular Educational Technologies 
(EdTech) \cite{naqvi2015data}. 
%
Currently, one of the grand challenges facing 
higher education is the problem of 
student retention and graduation \cite{parker2015advising}. 
There is a critical need to develop new  EdTech applications 
that analyze the data collected by  universities to ensure that 
students graduate in a timely fashion (4 to 6 years), and they are well prepared 
for jobs in their respective fields 
of study. 
To this end, several 
universities deploy a suite of software and tools.
For example,
\emph{degree planners} \footnote{http://www.blackboard.com/mobile-learning/planner.aspx} 
assist students in deciding their majors or fields of study, choosing the 
sequence of courses 
within their chosen major and 
providing advice for achieving career 
and learning objectives. \emph{Early warning systems} \cite{simons2011national} inform advisors/students
of 
progress,  and additionally provide cues for 
intervention 
when students are at the risk of failing one or more
courses and dropping out of their program of study.
In this work, we focus on the problem of 
next-term grade prediction where the 
goal 
is to predict the grade that
a student is expected to 
obtain in a course that he/she may enroll in the next term (future). 
%
%

In the past few years, several algorithms have been 
developed to analyze educational data, including 
Matrix Factorization (MF) algorithms inspired from recommender system
research. MF methods  decompose 
the student-course (or student-task) grade matrix into 
two low-rank matrices, and  then the prediction of the grade for a student on an untaken course is calculated as the
product of the corresponding vectors in the two 
decomposed matrices \cite{pero2015comparison,hwang2015unified}. Traditional MF algorithms  have shown a 
strong ability to deal with sparse datasets \cite{koren2009matrix} and their extensions have 
incorporated temporal and dynamic information \cite{ju2015using}. In our 
setting, we consider that a student's knowledge is continuously being enriched while taking  a sequence of courses;
and
it is important to incorporate 
this dynamic influence of sequential courses within our models. 
Therefore, we present a novel approach referred as 
Matrix Factorization with Temporal Course-wise 
Influence (MFTCI) model to predict next term student 
grades. 
MFTCI considers that a student's grade  on a certain course is 
determined by two components: (i) the student's competence with respect to each course's topics, 
content and requirement, etc., and (ii)  student's previous 
performance over other courses. 
%
We performed a comprehensive set of experiments on 
various datasets. 
The 
experimental results show that the proposed method outperforms several state-of-the-art methods.  The main 
contributions of our work in this paper are as follows:
\begin{enumerate}
\item We model and incorporate temporal course-wise influence in addition to matrix factorization for
grade prediction. Our experimental results demonstrate significant improvement from course-wise influence.
\item Our model successfully captures meaningful course-wise influences which correlate to the course content.
\item The learned influences between pairs of courses help in understanding pre-requisite structures within programs and tuning academic program chains.
\end{enumerate}






\section{Related Work}






Over the past few years, several methods have been developed to model
student behavior and academic performance ~\cite{baker2010data,he2013examining}, and they 
gain improvement of learning outcomes \cite{edmSurvey}. %
%
%
Methods  influenced by 
Recommender System (RS) research~\cite{Aggarwal2016}, 
including Collaborative Filtering (CF)~\cite{Ning2015} 
and Matrix Factorization ~\cite{Koren2009},  
have attracted increasing attention in educational mining
applications which relate to student 
grade prediction \cite{thai2010recommender} and 
in-class assessment prediction \cite{elbadrawy2014personalized}. 
Sweeney \emph{et. al.}~\cite{Mack1,Mack2} performed an extensive study of several 
recommender system approaches including SVD, SVD-kNN and Factorization Machine (FM) 
to predict next-term grade performance. 
Inspired by content-based recommendation~\cite{Pazzani2007} approaches, 
Polyzou \emph{et. al.}~\cite{Grade2016} addressed the future course grade prediction 
problem with three approaches: course-specific regression, student-specific regression 
and course-specific matrix factorization. 
Moreover, neighborhood-based CF approaches~\cite{ray2011collaborative,bydvzovska2015collaborative,denley2013course} predict grades 
based on the student similarities, i.e., they first identify similar students and use their 
grades to estimate the grades of the students with similar profiles.

In order to capture the changing of user dynamics over time in RS,  
various dynamic models have been developed. Many of such models are based on 
Matrix Factorization and state space models. 
Sun \emph{et. al.}~\cite{sun2014collaborative,sun2012dynamic} 
model user preference change using a state space model on latent user factors, and 
estimate user factors over time using noncausal Kalman filters. 
Similarly, Chua \emph{et.al.}~\cite{chua2013modeling} 
apply Linear Dynamical Systems (LDS) on Non-negative Matrix Factorization (NMF) to model user 
dynamics. 
Ju \emph{et. al.}~\cite{ju2015using} 
encapsulate the temporal relationships within a Non-negative matrix formulation. 
%
%
Zhang \emph{et. al.}~\cite{zhang2014latent} learn an explicit transition matrix over the 
latent factors for 
each user, and estimate the user and item latent factors and the transition matrices 
within a Bayesian framework.  
Other popular methods for dynamic modeling include time-weighting similarity decaying~\cite{Ding2005}, 
tensor factorization~\cite{Xiong2010} and point processes~\cite{Luo2015}. 
The method proposed in this paper tackle the challenges of next-term grade prediction which relates to 
the evolvement of student knowledge over taking a sequence of courses. 
Our key contribution involves how we incorporate the  temporal course-wise relationships within a MF approach. Additionally, the proposed 
approach learns pairwise relationships between courses that can help in understanding pre-requisite structures within programs and tuning 
academic program chains. 

\section{Preliminaries}
\subsection{Problem Statement and Notations}

Formally, student-course grades will be represented by a series 
of matrices \{G$_1$, G$_2$, ..., G$_T$\}  for $T$ terms. Each row 
of $G_t$ represents a student, each column of $G_t$ represents a course,
  and each value in $G_t$, denoted as $g_{s, c}^t$, represents a grade that student $s$ got on
  course $c$
  in term $t$ ($g_{s, c}^t \in (0, 4]$, $g_{s, c}^t=0$ indicates that 
  student $s$ did not take the course $c$ in
term $t$. We add a small value to failing grade to distinguish 0 score from such situation.). Student-course 
grades 
up to the $t_{th}$  term will be represented by G$^t$=$\sum_{i=1}^{t}$G$_i$  with size of $n \times m$, 
where $n$ is the number of students and 
$m$ is the number of courses. 
%
%
Given the database of (student, course, grade) up to term $(T-1)$ (i.e., G$^{T-1}$), the 
next-term grade prediction problem 
is to predict grades for each student on courses they might enroll in the next term  $T$. 
To simplify the notations, if not specifically stated in this paper, we will use  $g_{s,c}$ to denote  $g^{t}_{s,c}$. Our 
testing set is then
(student, course, grade) triples in the $T_{th}$ term, represented by matrix G$_T$. 
Rows from the grade matrices 
representing a student $s$ will simply be represented as  $G(s,:)$ and the specific courses that student has a grade for in this 
row can be given by $c' \in G(s,:)$. 

%

In this paper, all vectors (e.g., $\mathbf{u}_s^{\mathsf{T}}$ and $\mathbf{v}_c$) are represented by bold lower-case letters and all matrices (e.g., $A$) are represented 
by upper-case letters.  Column vectors are represented by having the transpose supscript$^{\mathsf{T}}$, otherwise by default they are 
row vectors. A predicted/approximated value is denoted by having a $\tilde \ $ head.




\section{Methods}
\subsection{MF with Temporal Course-wise Influence}
We consider the student $s$' grade on a certain course $c$, denoted as $g_{s, c}$, as determined by 
two factors. 
The first factor is the student $s$' competence with respect to the course $c$'s topics, content
and requirement. This is modeled through a latent factor model, in which $s$' competence is 
captured using a size-$k$ latent factor $\mathbf{u}_s$, $c$'s topics and contents are captured using a size-$k$
latent factor $\mathbf{v}_c$ in the same latent space as $\mathbf{u}_s$. 
Then the 
competence of $s$ over $c$ is modeled by the ``similarity'' between 
$\mathbf{u}_s$ and $\mathbf{v}_c$ via their dot product (i.e., $\mathbf{u}_s^{\mathsf{T}} \mathbf{v}_c$).

The second factor is the previous performance of student $s$ over other courses. We hypothesize that 
if course $c^{\prime}$ has a positive influence on course $c$, and student $s$ achieved 
  a high grade on 
$c^{\prime}$, then $s$ tends to have a high grade on $c$. 
Under this hypothesis, we model this second factor as 
a product between  the performance of student on a previous ``related'' course where the pairwise course relationships
are learned in our formulation. 
%
%
Note that we consider this pairwise course
influence as time independent, i.e., 
the influence of one course over another 
does  not change over time. However, the impact from previous performance/grades can be modeled using a 
decay function over time.
Taking these two factors, the estimated grade is given  as follows:
\begin{equation}
\label{eq:grade}
\begin{aligned}
 \tilde g_{s,c} & =  \mathbf{u}_s^{\mathsf{T}} \mathbf{v}_c \\
                & +   \underbrace{e^{-\alpha}\frac{\sum_{c^{\prime} \in G_{T-1}(s,:)} A(c^{\prime}, c)g_{s, c^{\prime}}}{|G_{T-1}(s,:)|}}_{\Delta (T-1)} \\
		& + \underbrace{e^{-2\alpha}\frac{\sum_{c^{\prime\prime} \in G_{T-2}(s, :)} A(c^{\prime\prime}, c) g_{s, c^{\prime\prime}}}{|G_{T-2}(s, :)|}}_{\Delta (T-2)},\\
\end{aligned}
\end{equation}
in which $A(c^{\prime}, c)$ is the influence of $c^{\prime}$ on $c$,
 $G_{T-1}(s, :)$/$G_{T-2}(s, :)$ is the subset of courses out of all courses  
 that $s$ has taken in the first/second previous terms, $|G_{T-1}(s, :)|$/$|G_{T-2}(s, :)|$ is the number of such taken courses.
$e^{-\alpha}$/$e^{-2\alpha}$ denote the time-decay factors. 
In Equation~\ref{eq:grade}, we consider previous two terms. More previous terms can 
be included with even stronger time-decay factors. 
%
Given the grade estimation as in Equation~\ref{eq:grade}, we formulate the grade prediction problem for term $T$ as 
the following optimization problem, 
\begin{equation}
\label{eq:opt}
\begin{aligned}
\underset{U, V, A}{\min}\ & \frac{1}{2}\sum_{s, c} (g_{s,c} - \tilde g_{s, c})^2 
 + \frac{\gamma}{2}(\|U\|_F^2 + \|V\|_F^2)\\
 &     + \tau \|A\|_* + \lambda\|A\|_{\ell_1}\nonumber\\
 & \text{s.t., }  A \ge 0
 \end{aligned}
\end{equation}
%
where $U$ and $V$ are the latent non-negative student factors and course factors, respectively; $\|A\|_*$ is the 
nuclear norm of $A$, which will induce an $A$ of low rank; and $\|A\|_{\ell_1}$ is the $\ell_1$ norm of $A$, 
which will introduce sparsity in $A$. 
In addition, the non-negativity constraint on $A$ is to enforce only positive influence across courses. 
\\[-11pt]
\subsubsection{Optimization Algorithm of MFTCI}
We apply the ADMM \cite{boyd2011distributed} technique for Equation \ref{eq:opt} by reformulating 
the optimization problem 
as follows, 
\begin{eqnarray*}
\label{eq:opt2}
& \underset{U, V, A, U_1, U_2, Z_1, Z_2}{\min}\ & \frac{1}{2}\sum_{s, c} (g_{s,c} - \tilde g_{s, c})^2+ \frac{\gamma}{2}(\|U\|_F^2 + \|V\|_F^2)\\
& &+ \tau \|Z_1\|_*+ \lambda\|Z_2\|_{\ell_1}\nonumber\\
& &+ \frac{\rho}{2} (\|A - Z_1\|_F^2 + \|A - Z_2\|_F^2)\nonumber \\
&  & + \rho (tr(U_1^{\mathsf{T}} (A - Z_1)))\\
&  &+ \rho (tr(U_2^{\mathsf{T}}(A - Z_2))) \nonumber\\
& \text{s.t., } & A \ge 0
\end{eqnarray*}
where $Z_1$ and $Z_2$ are two auxiliary variables, and $U_1$ and $U_2$ are two dual variables. 
All the variables are solved via an alternating approach as follows. 
\\[-20pt]
\paragraph{\underline{Step 1: Update $U$ and $V$}}
Fixing all the other variables and solving for $U$ and $V$, the problem becomes a classical matrix factorization 
problem: 
\begin{equation}
\label{eq:opt:uv}
\underset{U, V}{\min} \ \frac{1}{2}\sum_{s, c} (f_{s, c} - \mathbf{u}_s^{\mathsf{T}} \mathbf{v}_c)^2 + \frac{\gamma}{2}(\sum_s \|u_s\|_2^2 + \sum_c\|v_c\|_2^2)
\end{equation}
where $f_{s, c} = g_{s, c} - \Delta (T-1) - \Delta (T-2)$ (See Eq \ref{eq:grade}).
The matrix factorization problem can be solved using alternating minimization. 
\\[-20pt]
\paragraph{\underline{Step 2: Update $A$}}
Fixing all the other variables and solving for $A$, the problem becomes
\begin{eqnarray*}
\label{eq:opt:A}
& \underset{A}{\min}\  & \frac{1}{2}\sum_{s, c} (g_{s,c} - \tilde g_{s, c})^2 + \frac{\rho}{2} (\|A - Z_1\|_F^2 + \|A - Z_2\|_F^2) \\
&                                                     & + \rho (tr(U_1^{\mathsf{T}} (A - Z_1))) + \rho (tr(U_2^{\mathsf{T}}(A - Z_2))) \\
& \text{s.t., } & A \ge 0
\end{eqnarray*}
Using the gradient descent, the elements in $A$ can be updated as follows. 
\begin{equation}
\label{eq:opt:A:final}
\begin{aligned}
 A & (c_i, c_j)  =  A(c_i, c_j) - lr \times [\rho(A(c_i, c_j) - Z_1(c_i, c_j)) \\
&+ \rho (A(c_i, c_j) - Z_2(c_i, c_j)) + \rho U_1(c_i, c_j) + \rho U_2(c_i, c_j) \\
& -  \sum_{s, c_j} (g_{s, c_j} - \tilde g_{s, c_j})\\
&\times
\begin{cases}
\frac{e^{-\alpha}}{|G_{T-1}(s,:)|} g_{s, c_i} \quad \text{(if $c_i$ is taken in term $T-1$)} \\
\frac{e^{-2\alpha}}{|G_{T-2}(s,:)|} g_{s, c_i}  \quad \text{(if $c_i$ is taken in term $T-2$)}]
\end{cases}
\end{aligned}
\end{equation}
with projection into $[0, +\infty)$, where $lr$ is a learning rate. 
\\[-20pt]
\paragraph{\underline{Step 3: Update $Z_1$ and $Z_2$}}
For $Z_1$, the problem becomes
%
\begin{equation}
\label{eq:opt:Z1}
\begin{aligned}
\underset{Z_1}{\min}\ \tau \|Z_1\|_* + \frac{\rho}{2} \|A - Z_1\|_F^2 + \rho (tr(U_1^{\mathsf{T}}(A - Z_1)))
\end{aligned}
\end{equation}
%
The closed-form solution of this problem is 
\begin{equation}
Z_1 = S_{\frac{\tau}{\rho}}(A + U_1)
\end{equation}
where $S_{\alpha}(X)$ is a soft-thresholding function that shrinks the singular values of $X$ with a 
threshold $\alpha$, that is, 
\begin{equation}
S_{\alpha}(X) = U \text{diag}((\Sigma - \alpha)_+)V^{\mathsf{T}}
\end{equation}
where $X = U \Sigma V^{\mathsf{T}}$ is the singular value decomposition of $X$, and 
\begin{equation}
\label{eqn:max}
(x)_+ = \max(x, 0).
\end{equation}
For $Z_2$, the problem becomes
\begin{eqnarray}
\label{eq:opt:Z2}
\begin{aligned}
\underset{Z_2}{\min}\ \lambda \|Z_2\|_{\ell_1} + \frac{\rho}{2} \|A - Z_2\|_F^2 + \rho(tr(U_2^{\mathsf{T}})(A - Z_2))
\end{aligned}
\end{eqnarray}
The closed-form solution is 
\begin{equation}
	Z_2 = E_{\frac{\lambda}{\rho}}(A + U_2)
\end{equation}
where $E_{\alpha}(X)$ is a soft-thresholding function that shrinks the values in $X$ with a threshold 
$\alpha$, that is, 
\begin{equation}
	E_{\alpha}(X) = (X - \alpha, 0)_{+} 
\end{equation}
where $()_+$ is defined as in Equation~\ref{eqn:max}. 
\\[-20pt]
\paragraph{\underline{Step 4: Update $U_1$ and $U_2$}}
$U_1$ and $U_2$ are updated based on standard ADMM updates:
\begin{equation}
\label{eqn:updateU1}
	U_1 = U_1 + (A - Z_1); \quad\quad U_2 = U_2 + (A - Z_2)
\end{equation}

In addition, we conduct computational complexity analysis of MFTCI and put it in Appendix.

\section{Experiments}
\subsection{Dataset Description}
We evaluated our method on student grade 
records obtained from 
George Mason University (GMU) from Fall 2009 to Spring 2016. 
This period included data 
for 23,013 transfer students and 
20,086 first-time freshmen (non-transfer i.e., students who 
begin their study at  GMU) 
across 151 majors enrolled in 4,654 courses. 

Specifically, we extracted data for
six large and diverse majors for both 
non-transfer and transfer students. These majors 
include:  
(i) Applied Information Technology  (AIT), (ii) Biology (BIOL), (iii) Civil, Environmental and Infrastructure Engineering (CEIE), (iv) Computer Engineering (CPE)   (v) Computer Science (CS)   and (vi) Psychology (PSYC). 
Table  \ref{tab:statics} provides more 
information 
about  these 
datasets. 

\begin{table}
\caption{Dataset Descriptions}~\label{tab:statics}
\centering
\begin{threeparttable}
\begin{tabular}{c| c c c | c c c}
\hline
 \multirow{2}{*}{Major} & \multicolumn{3}{ c|}{Non-Transfer Students} & \multicolumn{3}{c}{Transfer Students} \\
\cline{2-7}
& \#S &  \#C & \#(S,C) & \#S & \#C &  \#(S,C)\\
\hline
AIT & 239 & 453 &  5,739  & 982 & 465 &  14,396 \\

BIOL & 1,448 & 990 & 33,527   & 1,330 & 833 &  22,691 \\

CEIE & 393  & 642 &  9,812  & 227 & 305 &  4,538 \\

CPE &  340  & 649 &  7,710  & 91  & 219 &  1,614 \\

CS & 908 & 818 & 18,376 &  480 & 464  &  7,967 \\

PSYC  & 911  & 874 & 22,598 & 1504 & 788 & 24,661 \\

\hline

Total & 4,239 & 1,115  &  97,762  & 4,614  & 1,019 & 75,867  \\

\hline
\end{tabular}

\begin{tablenotes}[para,flushleft]
\scriptsize{\#S, \#C and \#S-C are  number of students, courses and student-course pairs in educational records across the 6 majors from Fall 2009 to Spring 2016, respectively.}
\end{tablenotes} 

\medskip
\end{threeparttable} 
\end{table}











\subsection{Experimental Protocol}

\begin{figure}
\centering
  \includegraphics[width=1\linewidth]{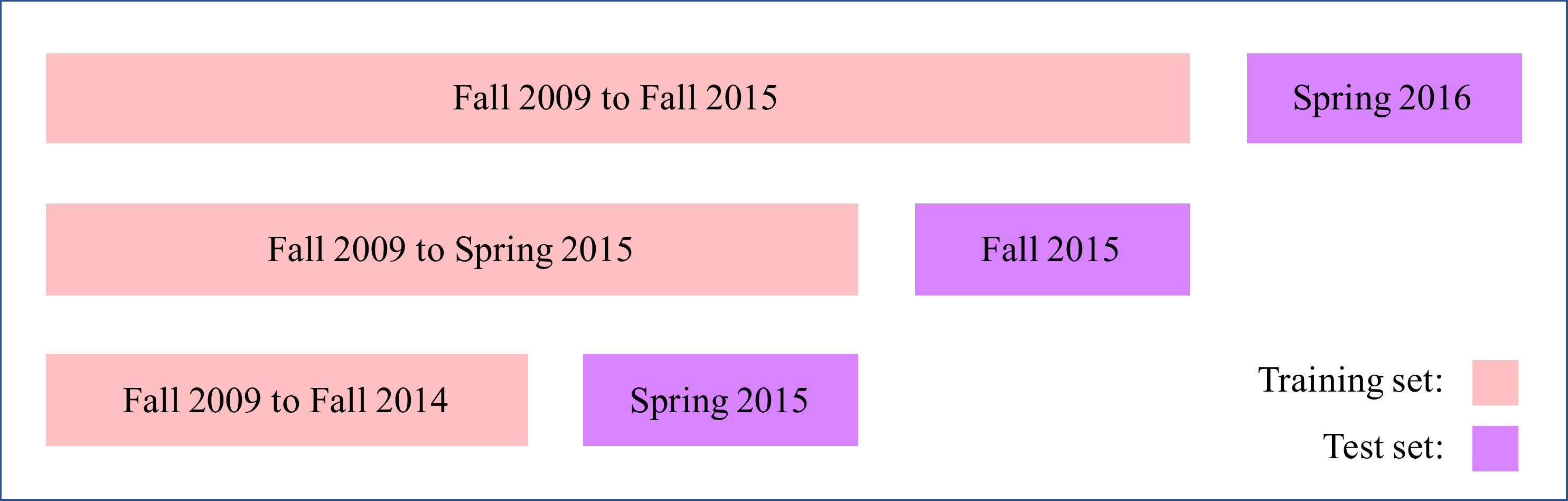}
  \caption{Different Experimental Protocols}~\label{fig:graph1}\
\end{figure}

To assess the performance of our next-term grade prediction models, we trained our models on data up to term $T-1$ and make predictions 
for term $T$. We evaluate our method for three 
  test terms, i.e., Spring 2016, Fall 2015 and Spring 2015.  As an 
  example, for evaluating predictions for term 
  Fall 2015, data from Fall 2009 to Spring 2015 is 
 considered as training data and data from Fall 2015 is testing data. 
%
%
datasets. Figure \ref{fig:graph1} shows the three different train-test splits.

\subsection{Evaluation Metrics}

We use
\textbf{Root Mean Squared Error (RMSE)} and \textbf{Mean Absolute Error (MAE)}  as 
metrics for evaluation, and are defined as follows:
\begin{equation*}
\begin{aligned}
\begin{split}
RMSE = \sqrt{\frac{\sum_{s,c\in G_T}(g_{s,c} - \tilde{g}_{s,c})^2}{\left | G_T  \right |}}, \\
MAE = \frac{\sum_{s,c\in G_T}\left |g_{s,c} - \tilde{g}_{s,c} \right |}{\left | G_T  \right |}
\end{split}
\end{aligned}
\end{equation*}
where $g_{s,c}$ and $\tilde{g}_{s,c}$ are the ground truth and predicted grade for student $s$ on course $c$, and $G_T$ is the testing set of (student, course, grade) triples in the $T_{th}$ term. Normally, in next-term grade prediction problem, MAE is more intuitive than RMSE since MAE is a straightforward method which calculates the deviation of errors directly while RMSE has implications such as penalizing large errors more. 

For our dataset, a student's grade can be a letter grade (i.e.  A, A-, \ldots, F). As done previously 
by Polyzou et. al. \cite{polyzou2016grade} we define
a tick to denote the difference between two consecutive letter grades (e.g., C+ vs C or C vs C-). To assess
the performance of our grade prediction method, we convert
the predicted grades into their closest letter grades and compute the percentage of 
predicted grades with no error (or 0-ticks), within 1-tick and within 2-ticks denoted
by Pct$_0$, Pct$_1$ and Pct$_2$, respectively. For the problem of course selection and degree 
planning, courses predicted within 2 ticks can be considered sufficiently correct. We name 
these metrics as \textbf{Percentage of Tick Accuracy (PTA)}.

\subsection{Baseline Methods}

We compare the performance of our proposed method to the following baseline approaches.

\subsubsection{Matrix Factorization}

Matrix factorization is known to be successful in predicting ratings accurately in recommender systems ~\cite{ricci2011recommender}. This approach can be applied directly on next-term grade prediction problem by considering student-course grade matrix as a user-item rating matrix in recommender  systems. Based on the assumption that each course and student can be represented in  the 
same low-dimensional space, corresponding to the knowledge space,  two low-rank matrices containing latent factors are learned
to represent courses and students ~\cite{Mack2}. Specifically, the grade a student  $s$ will achieve on a course $c$ is predicted as follows:
 \begin{eqnarray}
\label{eq:mf}
\begin{aligned}
 \tilde g_{s,c}  =  \mu + \mathbf{p}_s +  \mathbf{q}_c + \mathbf{u}_s^{\mathsf{T}} \mathbf{v}_c
\end{aligned}
\end{eqnarray}
where $\mu$ is a global bias term, $\mathbf{p}_s$ ($\mathbf{p} \in \mathbb{R}^n$) and $\mathbf{q}_c$ ($\mathbf{q} \in \mathbb{R}^m$) are the student and course bias terms (in this case, for student $s$ and course $c$), respectively, and $\mathbf{u}_s$ ($\mathbf{U} \in \mathbb{R}^{k \times n}$) and $\mathbf{v}_c$ ($\mathbf{V} \in \mathbb{R}^{k \times m}$) are the latent factors for student $s$ and course $c$, respectively. 

\subsubsection{Matrix Factorization without Bias (MF$_0$)}
We only considered the student and course latent factors to predict the next-term grades. Therefore, the grade a student $s$ will achieve on a course $c$ is calculated as follows:
 \begin{eqnarray}
\label{eq:mfnoBias}
\begin{aligned}
 \tilde g_{s,c}  = \mathbf{u}_s^{\mathsf{T}} \mathbf{v}_c
\end{aligned}
\end{eqnarray}
 


\subsubsection{Non-negative Matrix Factorization (NMF) ~\cite{lee2001algorithms}} We add non-negative constraints
on matrix $\mathbf{U}$ and matrix $\mathbf{V}$ in Equation \ref{eq:mfnoBias}.
The non-negativity constraints allows MF approaches to have better interpretability and accuracy for  non-negative data  \cite{ho2008nonnegative}.

\section{Results and Discussion}

\subsection{Overall Performance}
\begin{table*}[th!]
\caption{Comparison Performance  with PTA (\%)}~\label{tab:tick}
\centering
\begin{threeparttable}
\begin{tabular}{
                @{\hspace{0pt}}r@{\hspace{3pt}}|
                @{\hspace{0pt}}r@{\hspace{3pt}}
                @{\hspace{3pt}}r@{\hspace{3pt}}
                @{\hspace{3pt}}r@{\hspace{3pt}}|
                @{\hspace{3pt}}r@{\hspace{3pt}}
                @{\hspace{3pt}}r@{\hspace{3pt}}
                @{\hspace{3pt}}r@{\hspace{3pt}}|
                @{\hspace{3pt}}r@{\hspace{3pt}}
                @{\hspace{3pt}}r@{\hspace{3pt}}
                @{\hspace{3pt}}r@{\hspace{3pt}}
}
\hline
 \multirow{2}{*}{Methods} &
 \multicolumn{3}{c| @{\hspace{3pt}}}{Spring 2016} &  \multicolumn{3}{c| @{\hspace{3pt}}}{Fall 2015} &  \multicolumn{3}{c @{\hspace{3pt}}}{Spring 2015} \\
\cline{2-10}

& {Pct$_0$($\uparrow$)}  & {Pct$_1$($\uparrow$)}  & {Pct$_2$($\uparrow$)}  &  {Pct$_0$} & {Pct$_1$}  & {Pct$_2$}  &  {Pct$_0$} & {Pct$_1$}  & {Pct$_2$} \\
\hline

MF &   {13.25}  & {27.71}   &  {58.02}  &  {12.05} &  {26.63} &   {58.89}&   {13.03}& {26.09}  &  {54.83} \\
MF$_0$ &   {16.52}  & {31.65}   &  {57.46}  &  {15.51} &  {30.03} &   {55.64}&   {15.53}& {29.53}  &  {54.94} \\
NMF &   {13.21}  & {27.04}   &    {57.18}& {15.33}  &  {30.12} &  {56.15} &    {15.56}& {29.23}  &  {54.93} \\
\hline
MFTCI  &   {{\bf 19.78}}  & {{\bf 35.52}}   &  {{\bf 61.44}}  &  {{\bf 19.71}} &  {{\bf 35.16}} &   {{\bf 60.12}}&   {{\bf 18.56}}& {{\bf 32.78}}  &  {{\bf 58.80}} \\
\hline

\end{tabular}

\begin{tablenotes}[para,flushleft]
\scriptsize{ i)  ``$\uparrow$" indicates the higher the better.
	ii) Reported values of Pct$_0$, Pct$_1$ and Pct$_2$ are percentages.
	 iii) Best performing methods are highlighted with bold. }
\end{tablenotes} 
\medskip
\end{threeparttable} 
\end{table*}
%
%
Table \ref{tab:tick} presents the comparison of Pct$_0$, Pct$_1$ and Pct$_2$ for
non-transfer students for the three terms considered as test:  Spring 2016, Fall 2015 and  Spring 2015. 
We observe that the MFTCI model outperforms the baselines across the different test sets. On 
average, MFTCI outperforms the MF, MF$_0$ and NMF methods by 34.18\%, 11.59\% and 4.08\% in terms of Pct$_0$,  16.64\%, 7.96\% and 4.03\% in terms of Pct$_1$,  and  2.10\%, 3.00\% and 1.98\% in terms of Pct$_2$, respectively. We observe similar 
results for transfer students as well (not included here for brevity).

Table \ref{tab:predPastTerm} presents the performance of the baselines and MFTCI model 
for the three different terms of both non-transfer and transfer students
  using RMSE and MAE  as evaluation metrics. The MFTCI model  consistently outperforms  the  baselines across the different datasets in terms of MAE.
In addition, the results shows that MF$_0$, NMF and MFTCI tend to have better performance for Spring 2016 term than Fall 2015 term.
Similar trend is observed between Fall 2015 term and Spring 2015 term. This suggests that MFTCI is likely to have better performance with more information in the training set.

\begin{table*}[th!]
\caption{Comparison Performance  with RMSE and MAE.}~\label{tab:predPastTerm}
\centering
\begin{tabular}{
                @{\hspace{0pt}}r@{\hspace{3pt}}|
                @{\hspace{0pt}}r@{\hspace{3pt}}
                @{\hspace{3pt}}r@{\hspace{3pt}}
                @{\hspace{3pt}}r@{\hspace{3pt}}
                @{\hspace{3pt}}r@{\hspace{3pt}}
                @{\hspace{3pt}}r@{\hspace{3pt}}
                @{\hspace{3pt}}r@{\hspace{3pt}}
                @{\hspace{3pt}}r@{\hspace{3pt}}
                @{\hspace{3pt}}r@{\hspace{3pt}}
                @{\hspace{0pt}}r@{\hspace{0pt}}|
                @{\hspace{3pt}}r@{\hspace{3pt}}
                @{\hspace{3pt}}r@{\hspace{3pt}}
                @{\hspace{3pt}}r@{\hspace{3pt}}
                @{\hspace{3pt}}r@{\hspace{3pt}}
                @{\hspace{3pt}}r@{\hspace{3pt}}
                @{\hspace{3pt}}r@{\hspace{3pt}}
                @{\hspace{3pt}}r@{\hspace{3pt}}
                @{\hspace{3pt}}r@{\hspace{0pt}}
}
\hline
 \multirow{3}{*}{Methods} &
 \multicolumn{9}{c| @{\hspace{3pt}}}{Non-Transfer Students} &
 \multicolumn{8}{c }{Transfer Students} \\ 
 \cline{2-18}
 & \multicolumn{2}{c }{Spring 2016} & & \multicolumn{2}{c }{Fall 2015} & & \multicolumn{2}{c @{\hspace{3pt}}}{Spring 2015}& &
  \multicolumn{2}{c }{Spring 2016} & & \multicolumn{2}{c }{Fall 2015} & & \multicolumn{2}{c }{Spring 2015}  \\
 \cline{2-3}
 \cline{5-6} 
\cline{8-10}
\cline{11-12}
\cline{14-15}
\cline{17-18}
& {RMSE}  & {MAE} & & {$\ $RMSE$\ $}  & {$\ $MAE$\ $}  & & {$\ $RMSE$\ $} & {$\ $MAE$\ $}  &  &{$\ $RMSE$\ $}  & {$\ $MAE$\ $} &  & {$\ $RMSE$\ $} & {$\ $MAE$\ $} & & {$\ $RMSE$\ $} & {$\ $MAE$\ $}  \\
\hline

MF &   {0.999}  & {0.754}   & &   {1.037}  & {0.786}   & &  {1.023} & {0.784}  &   &   {0.925}  & {0.688}   & & {\bf 0.921}  &  {0.686}   & &{0.985}    &   {0.732}\\

MF$_0$ &   {0.929}  & {0.714}   & &   {0.977}  & {0.752}   & &  {1.014} & {0.778}  &   &   {0.893}  & {0.668}   & & {0.944}  &  {0.705}   & &{1.011}    &   {0.765}\\

NMF &   {1.020}  & {0.769}   & &   {\bf 0.967}  & {0.746}   & &  {\bf 1.000} & {0.771}  &   &   {0.906}  & {0.683}   & & {0.932}  &  {0.701}   & &{\bf 0.979}    &   {0.746}\\

\hline

MFTCI  &   {{\bf 0.928}}  & {{\bf 0.685}}   & &   {0.982}  & {{\bf 0.717}}   & &  {1.012} & {{\bf 0.750}}  &   &   {\bf 0.887}  & {{\bf 0.636}}   & & {0.927}  &  {{\bf 0.662}}   & &{1.000}    &   {{\bf 0.721}}\\

\hline
\end{tabular}
\end{table*}

\subsection{Analysis on Individual Majors}
\begin{table}[th!]
    \caption{Comparison Performance for Different Majors \label{tab:stdGrpMAE1}}
      \centering
            \begin{tabular}{
                @{\hspace{0pt}}r@{\hspace{0pt}}|
                @{\hspace{0pt}}r@{\hspace{0pt}}|
                @{\hspace{3pt}}r@{\hspace{3pt}}
                @{\hspace{3pt}}r@{\hspace{3pt}}
                @{\hspace{3pt}}r@{\hspace{3pt}}
                @{\hspace{3pt}}r@{\hspace{3pt}}
                @{\hspace{3pt}}r@{\hspace{3pt}}
                @{\hspace{3pt}}r@{\hspace{0pt}}
              }
             \hline

         &   {Methods}   &  {AIT} &   {BIOL} &   { CEIE} &   {CPE} &    { CS} &    {PSYC} \\
            \hline
   \multirow{4}{*}{{Pct$_0$}}  & {MF} & {18.71} & {18.00} & {15.99} & {12.99} & {15.98} & {20.18} \\
   							&{MF$_{0}$} & {19.45} & {22.10} & {16.70} & {14.21} & {16.47} & {22.12} \\
					 & {NMF} & {19.77} & {22.16} & {{\bf 17.01}} & {14.32} & {16.61} & {22.17} \\
					 & {MFTCI}& {{\bf 22.30}} & {{\bf 24.24}} & {16.80} & {{\bf 14.32}} & {{\bf 17.32}} & {{\bf 25.83}} \\
\hline				 
					 
   \multirow{4}{*}{{Pct$_1$}}  &  {MF} & {37.95} & {35.43} & {31.47} & {27.86} & {31.53} & {39.41} \\
							 &  {MF$_{0}$} & {37.21} & {39.68} & {31.87} & {{\bf 27.97}} & {30.51} & {39.63} \\
					& {NMF} & {36.79} & {39.74} & {31.67} & {27.19} & {30.43} & {39.36} \\
 					&  {MFTCI}& {{\bf 39.64}} & {{\bf 40.87}} & {{\bf 32.38}} & {27.53} & {{\bf 31.78}} & {{\bf 42.29}} \\
\hline

   \multirow{4}{*}{{Pct$_2$}} & {MF} & {{\bf 67.02}} & {67.78} & {58.66} & {52.28} & {56.91} & {{\bf 71.01}} \\
   							& {MF$_{0}$} & {66.17} & {67.54} & {58.35} & {50.72} & {56.24} & {67.74} \\
					& {NMF} & {66.70} & {67.54} & {58.55} & {51.17} & {56.17} & {67.79} \\
 					& {MFTCI}& {66.70} & {{\bf 68.25}} & {{\bf 58.76}} & {{\bf 52.94}} & {{\bf 58.18}} & {68.29} \\
\hline

        \end{tabular}
   \end{table}

We divide non-transfer students based on their majors and test the baselines and MFTCI model on each major, separately. Table \ref{tab:stdGrpMAE1} shows the comparison of Pct$_0$, Pct$_1$ and Pct$_2$ on different majors.  The results show that MFTCI has the best performance for
almost all the majors. Among all the results, MFTCI has the highest accuracy when predicting grades for PSYC and BIOL  students for which we have more student-course pairs in the training set.

\subsection{Effects from Previous Terms on MFTCI}
\begin{figure}[th!]
\centering
  \includegraphics[width=0.9\linewidth]{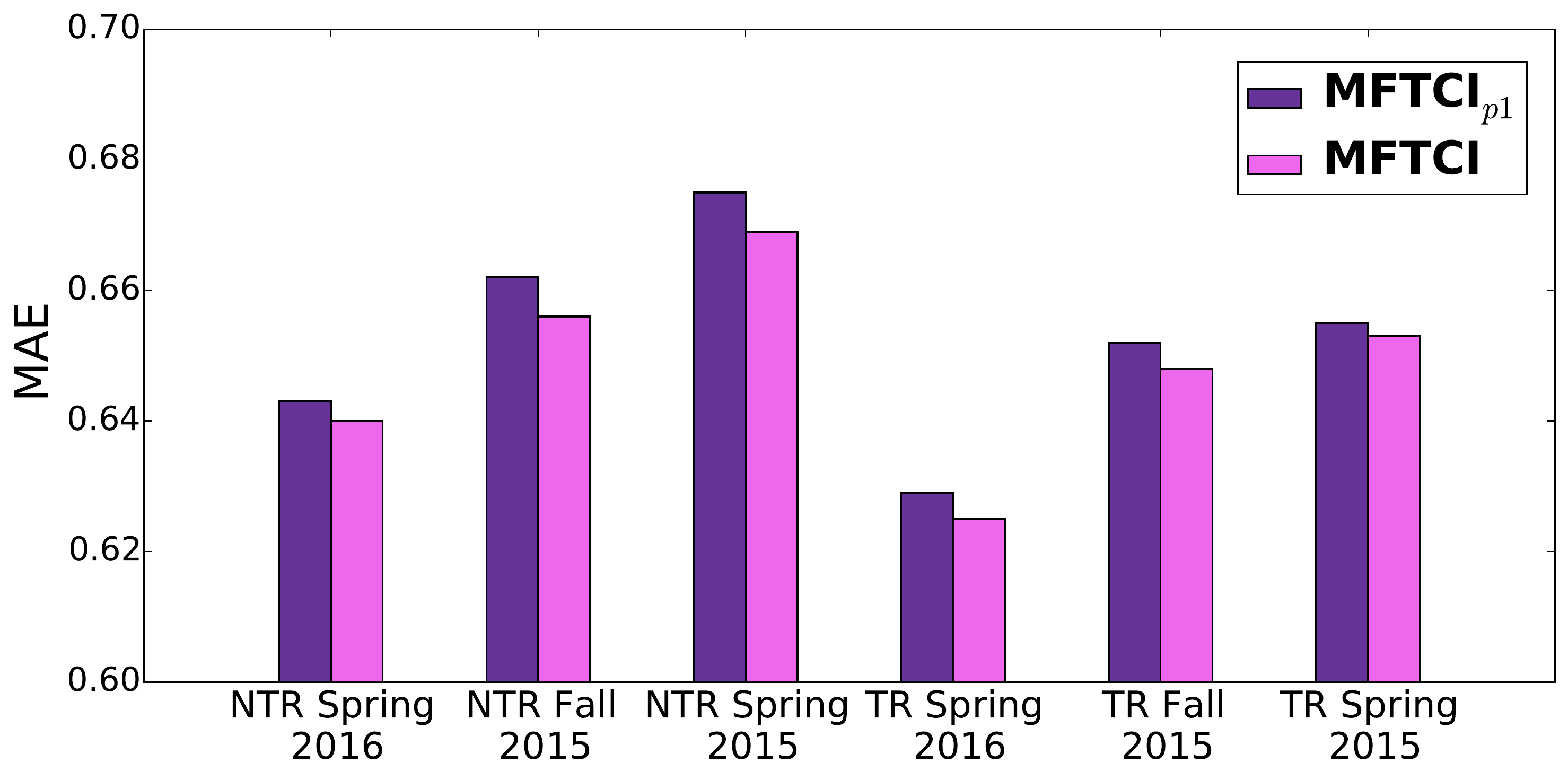}
  \caption{Comparison performance for MFTCI$_{p1}$ and MFTCI}~\label{fig:prev1}\
\end{figure}

In order to see the influence of number of previous terms considered in 
MFTCI,
we run our model with only $\Delta (T-1)$ in Equation \ref{eq:grade}. This method is 
represented as MFTCI$_{p1}$. 
Figure \ref{fig:prev1} shows the comparison results of MAE for six subsets of data
which are reported in Table \ref{tab:predPastTerm}, where ``NTR" stands for non-transfer students and ``TR" stands for transfer students. The results 
show that MFTCI consistently 
outperforms MFTCI$_{p1}$ on all datasets. This suggests that considering two previous terms is necessary for achieving good prediciton results. Moreover, since we consider that 
the student's knowledge is modeled using an exponential  decaying function over  time, we do not 
include the influence from the third previous term in our model as its influence for the 
grade prediction is negligible in comparison to  the previous two 
terms.

\subsection{Visualization of Course Influence}
\begin{figure*}[t]
\centering
  \includegraphics[width=0.9\linewidth]{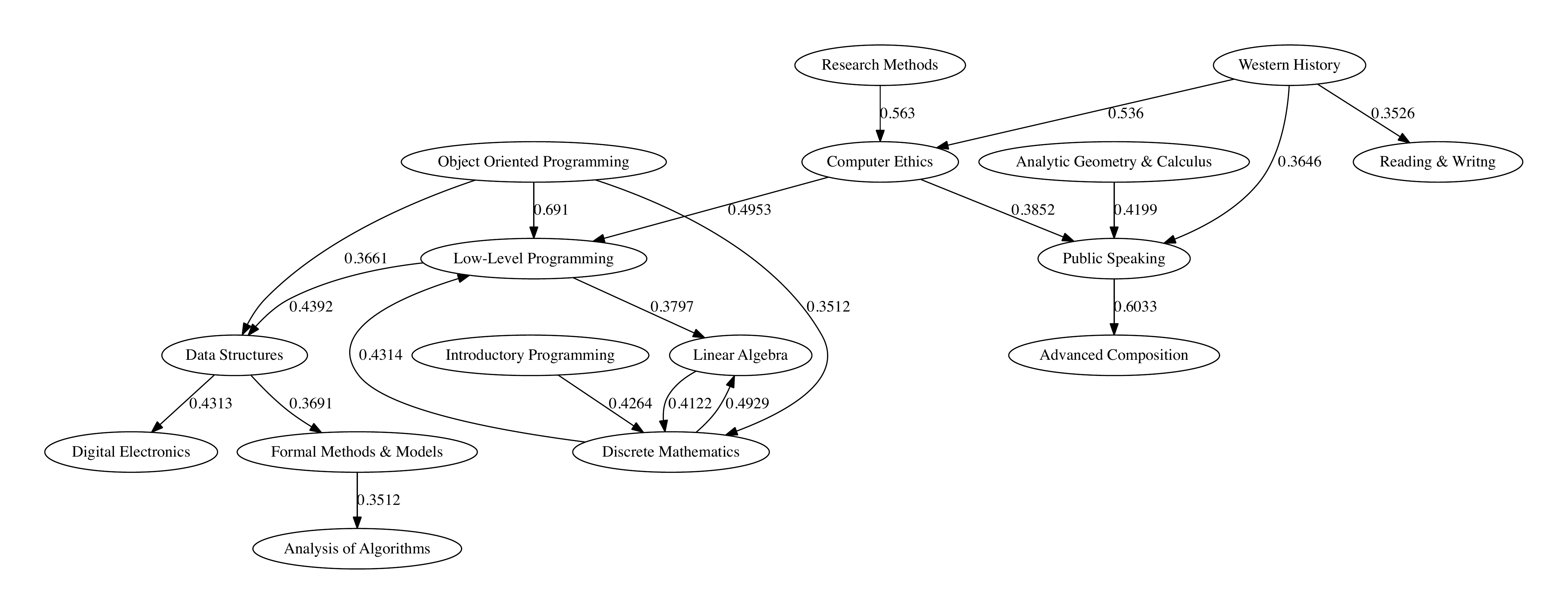}
  \caption{Identified course influences for CS major}~\label{fig:cstop20}\
\end{figure*}

\begin{figure*}[t]
\centering
  \includegraphics[width=0.9\linewidth]{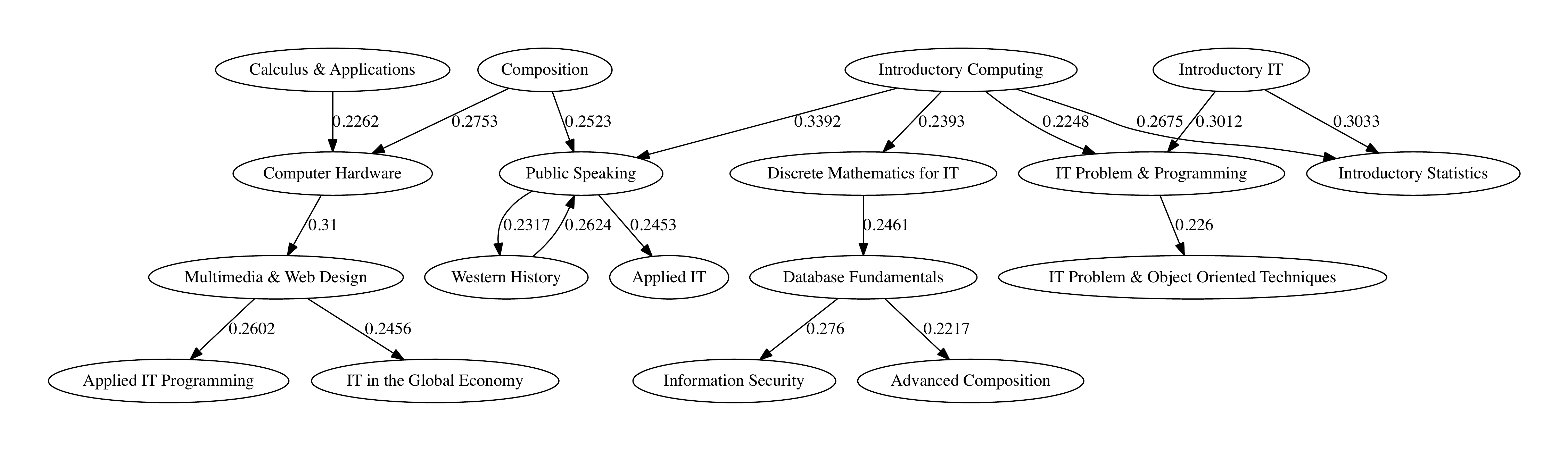}
  \caption{Identified course influences for AIT major}~\label{fig:aittop20}\
\end{figure*}
To  interpret what is captured in the course influence matrix $A$ (See Eq \ref{eq:grade}), 
we extract the top 20 values with the corresponding course names (and topics) 
for analysis.
Figure \ref{fig:cstop20} and \ref{fig:aittop20} show the captured pairwise course influences for CS and AIT majors, respectively. Each node
  corresponds to one course which is represented by the shortened course's name. We can notice from 
  the figures that most influences reflect content dependency between courses. For example, in 
 the  CS major, ``Object Oriented Programming" course has significant influence on  performance of 
 ``Low-Level Programming" course (the former one is also the latter one's prerequisite course); ``Linear Algebra" and ``Discrete Mathematics" have influence on each other; ``Formal Methods \& Models" course has influence on ``Analysis of Algorithms" course. 
 In case of the 
 AIT major, both ``Introductory IT" course and ``Introductory Computing" course have influence on ``IT Problem \& Programming" course; ``Multimedia \& Web Design" course has influence on both ``Applied IT Programming" course and ``IT in the Global Economy" course. GMU has a sample schedule 
 of eight-term courses for each major in order to guide undergraduate students to finish their study step by step based on the level, content and difficulty of courses \footnote{http://catalog.gmu.edu}. Among 
 the identified relationships 
 shown in Figures \ref{fig:cstop20} and \ref{fig:aittop20} we found 17 and  13 
 of the CS and AIT courses influences in the guide map, respectively.
 The rest of the identified influences  are among 
 other general electives but  required 
 courses (e.g., ``Public Speaking" course), or specific electives pertaining to the major (e.g., ``Research Methods'' course). This shows that our 
 model learns meaningful course-wise influences and
 successfully uses it to improve MF model. 
 


\begin{figure*}[t!]
        \centering
        \begin{subfigure}[b]{0.475\textwidth}
            \centering
            \includegraphics[width=\textwidth]{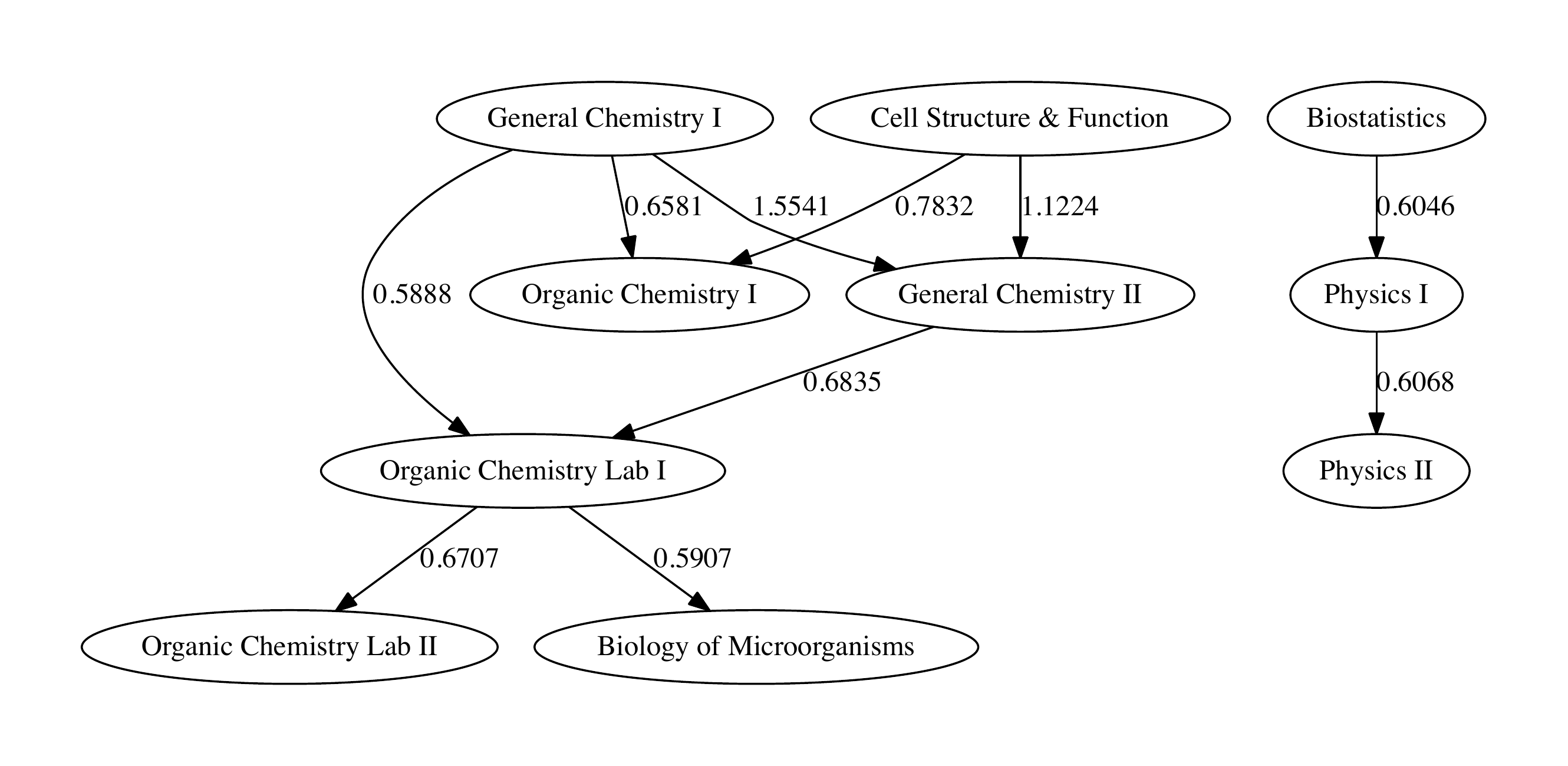}
            \caption{Identified course influences for BIOL major}~\label{fig:bioltop10}\
        \end{subfigure}
        \hfill
        \begin{subfigure}[b]{0.475\textwidth}  
            \centering 
            \includegraphics[width=\textwidth]{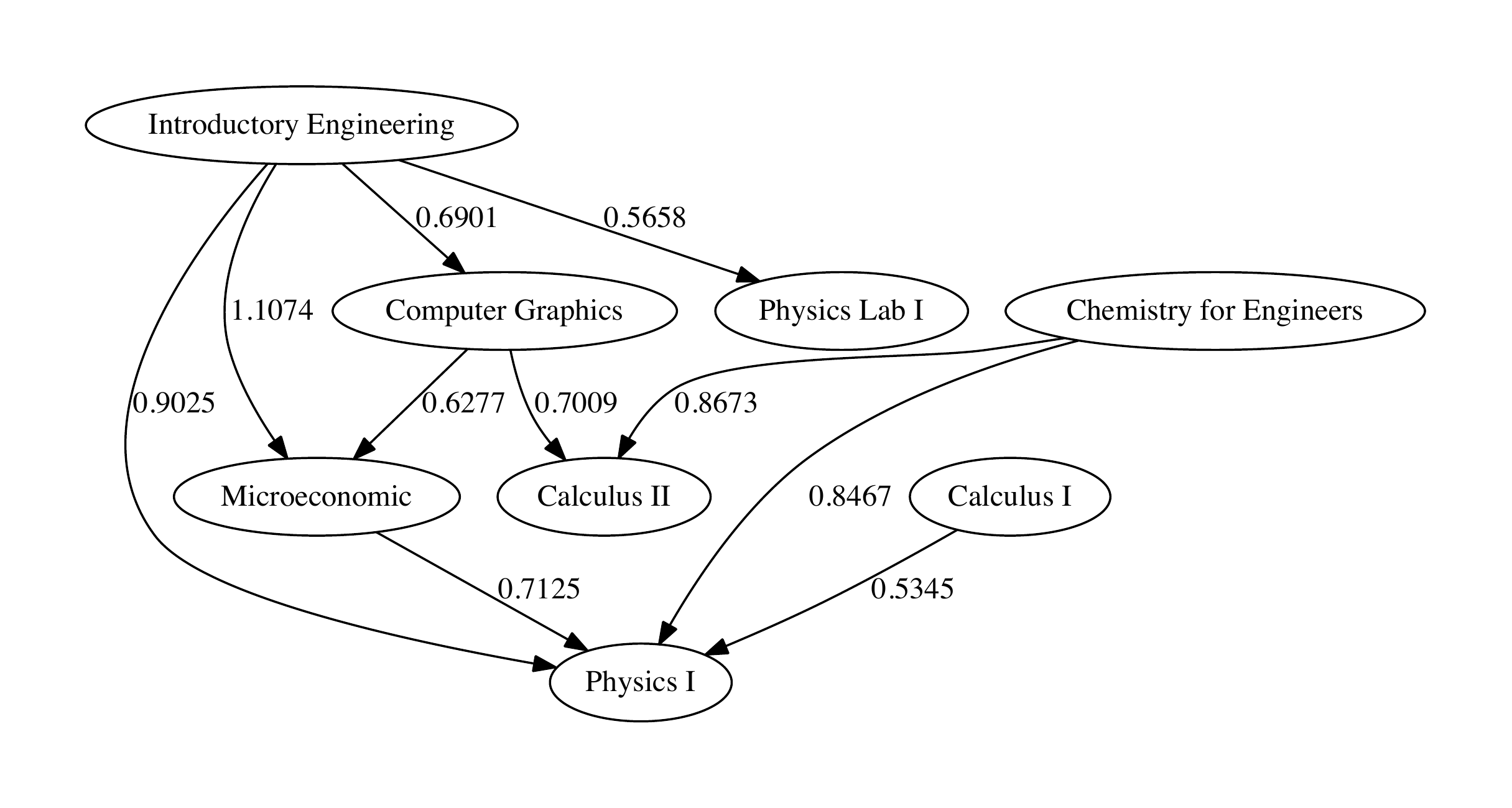}
             \caption{Identified course influences for CEIE major}~\label{fig:ceietop10}\
        \end{subfigure}
        \vskip\baselineskip
        \begin{subfigure}[b]{0.475\textwidth}   
            \centering 
            \includegraphics[width=\textwidth]{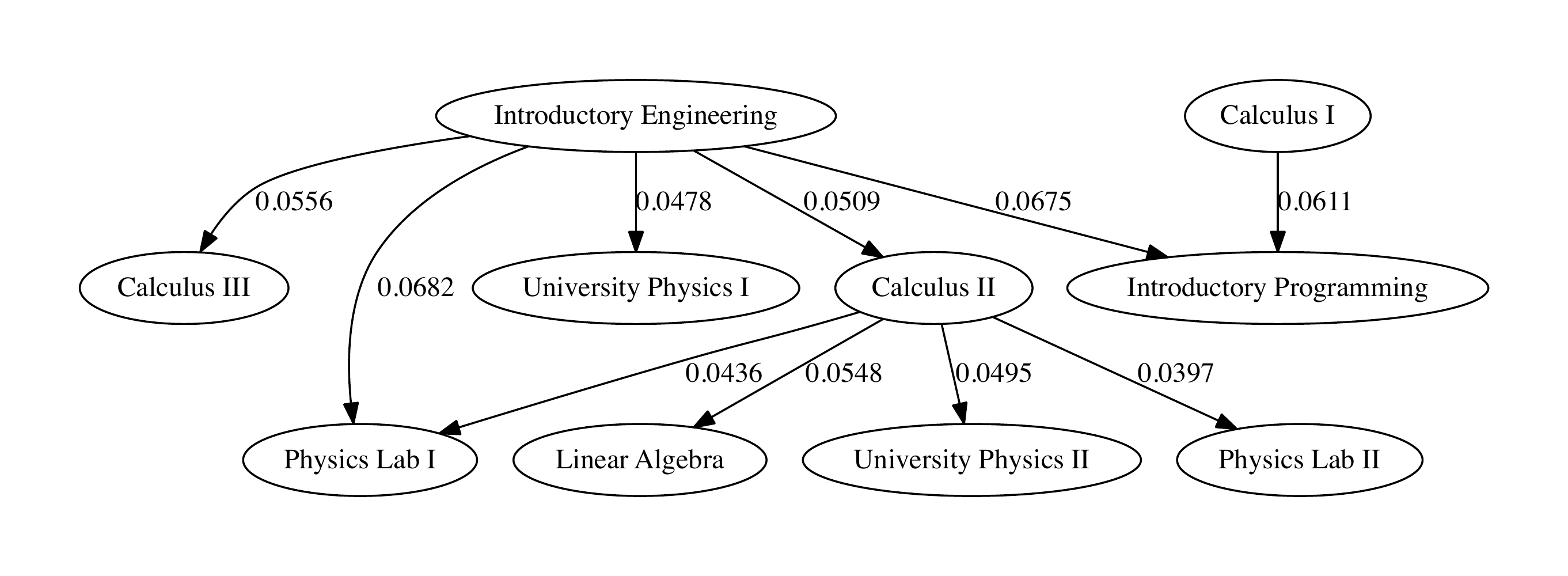}
            \caption{Identified course influences for CPE major}~\label{fig:cpetop10}\
        \end{subfigure}
        \quad
        \begin{subfigure}[b]{0.475\textwidth}   
            \centering 
            \includegraphics[width=\textwidth]{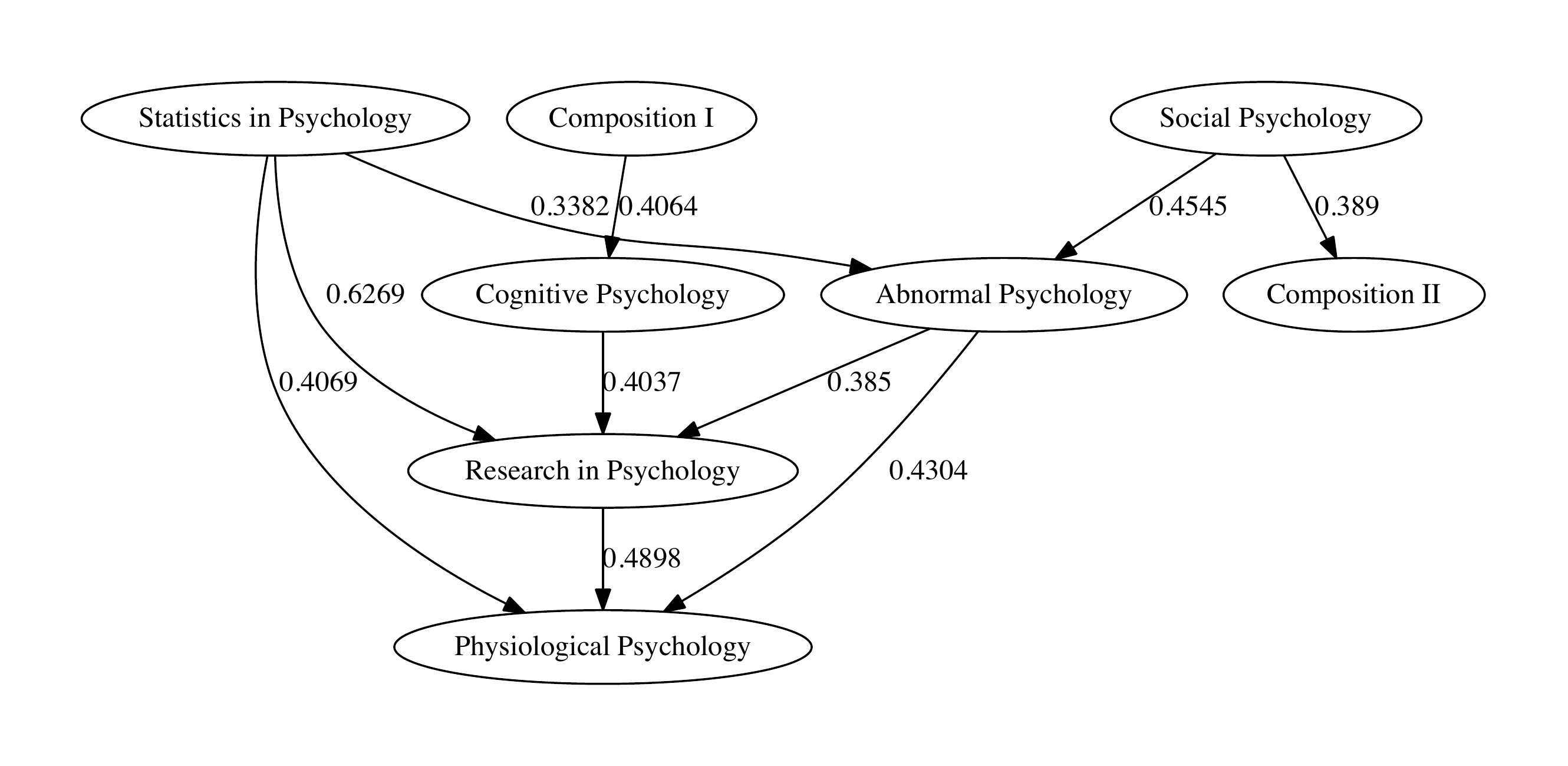}
            \caption{Identified course influences for PSYC major}~\label{fig:psyctop10}\
        \end{subfigure}
        \caption{Identified course influences for different majors}~\label{fig:majortop10}\
    \end{figure*}

Figure \ref{fig:majortop10} shows the identified course influences for the 
BIOL, CEIE, CPE and PSYC majors.
These identified course-wise influences seem to capture similarity of course content.
%


\section{Conclusion and Future Work}
We presented   a Matrix Factorization with Temporal Course-wise Influence (MFTCI) model 
that integrates 
factorization models and the influence of courses taken in the preceding terms to predict student 
grades for the next term. 

We evaluate our model on the student educational records from Fall 2009 to Spring 2016 collected from George Mason University. The dataset 
in this study contains 
both non-transfer and transfer students from six 
different majors. Our experimental evaluation shows that 
MFTCI consistently outperforms the different 
state-of-the-art methods.  Moreover, we analyze the effects from previous terms on MFTCI, and we make the conclusion that it is necessary to consider two previous terms. In addition, we visualize the patterns learned between pairs of courses.  The results strongly demonstrate
that the learned course 
influences  correlate with
the course content within academic programs.

In the future, we will explore incorporation of additional 
constraints over the the pairwise 
course influence matrix, such as prerequisite information, compulsory and elective 
provision of a course.  We will explore using the course influence information to build a degree planner for future students.



\section{Acknowledgments}
Funding was provided by NSF Grant, 1447489.

\appendix
\section{Computational Complexity Analysis}
The computational complexity of MFTCI is determined by the four steps in the alternating approach 
as described above. 
To update $U$ and $V$ as in Equation~\ref{eq:opt:uv} using gradient descent method via alternating 
minimization, the computational complexity is 
$O(\text{niter}_{uv}(k\times n_{s,c} + k\times m + k \times n))= O(\text{niter}_{uv}(k\times n_{s,c}))$
(typically $n_{s, c} \ge \max(m, n)$), where $n_{s,c}$ is the total number of 
student-course dyads, $n$ is the number of students, $m$ is the number of courses, $k$ is the latent 
dimensions of $U$ and $V$, and $\text{niter}_{uv}$ is the number of iterations. 
To update $A$ as in Equation~\ref{eq:opt:A:final} using gradient descent method, 
the computational complexity is upper-bounded by
$O(\text{niter}_a (n_{cc} \times \frac{n_{s, c}}{m}))$, where $n_{cc}$ is the number of course pairs that have been taken
by at least one student, $\frac{n_{s, c}}{m}$ is the average number of students for a course, which 
upper bounds the average number of students who co-take two courses, and 
 $\text{niter}_a$ is the number of iteractions. 
Essentially, to update $A$, we only need to update $A(c_i, c_j)$ where $c_i$ and $c_j$ have been 
co-taken by some students. For $A(c_i, c_j)$ where $c_i$ and $c_j$ have never been taken together, they 
will remain 0.
To update $Z_1$ as in Equation~\ref{eq:opt:Z1}, a singular value decomposition is involved and thus its 
computational complexity is upper bounded by $O(m^3)$. 
To update $Z_2$ as in Equation~\ref{eq:opt:Z2}, the computational
complexity is $O(m^2)$. 
To update $U_1$ and $U_2$ as in Equation~\ref{eqn:updateU1}, the computational complexity is 
$O(m^2)$. 
Thus, the computational complexity for MTFCI is 
$O(\text{niter}(\text{niter}_{uv}(k\times n_{s,c}) + \text{niter}_a (n_{cc} \times \frac{n_{s, c}}{m}) + m^3 + m^2))$ = $O(\text{niter}(\text{niter}_{uv}(k\times n_{s,c}) + \text{niter}_a (n_{cc} \times \frac{n_{s, c}}{m})+m^3))$, where $\text{niter}$ is the number of iterations for the four steps. Although 
the complexity is dominated by $m^3$ due to the 
SVD on $A+U_1$, since $n$ (i.e., the number of courses) is typically not large, the run time will be more 
dominated by $n_{s,c}$ (i.e., the number of student-course dyads).

\bibliography{mybib}{}
\bibliographystyle{plain}

\balancecolumns
\end{document}